\documentclass[conference]{IEEEtran}
\IEEEoverridecommandlockouts
\usepackage{cite}
\usepackage{amsmath,amssymb,amsfonts}
\usepackage{algorithmic}
\usepackage{graphicx}
\usepackage{textcomp}
\usepackage{xcolor}
\usepackage{adjustbox}
\usepackage{multirow}
\def\BibTeX{{\rm B\kern-.05em{\sc i\kern-.025em b}\kern-.08em
    T\kern-.1667em\lower.7ex\hbox{E}\kern-.125emX}}
\begin{document}

\title{Multi-level and Multi-modal Feature Fusion for Accurate 3D Object Detection in Connected and Automated Vehicles\\}

\author{\IEEEauthorblockN{Yiming Hou}
\IEEEauthorblockA{\textit{Institute for Transport Studies} \\
\textit{University of Leeds}\\
Leeds, UK \\
tsyh@leeds.ac.uk}
\and
\IEEEauthorblockN{Mahdi Rezaei}
\IEEEauthorblockA{\textit{Institute for Transport Studies} \\
\textit{University of Leeds}\\
Leeds, UK \\
m.rezaei@leeds.ac.uk}
\and
\IEEEauthorblockN{Richard Romano}
\IEEEauthorblockA{\textit{Institute for Transport Studies} \\
\textit{University of Leeds}\\
Leeds, UK \\
r.romano@leeds.ac.uk}
}

\maketitle

\begin{abstract}
Aiming at highly accurate object detection for connected and automated vehicles (CAVs), this paper presents a Deep Neural Network based 3D object detection model that leverages a three-stage feature extractor by developing a novel LIDAR-Camera fusion scheme. The proposed feature extractor extracts high-level features from two input sensory modalities and recovers the important features discarded during the convolutional process. The novel fusion scheme effectively fuses features across sensory modalities and convolutional layers to find the best representative global features. The fused features are shared by a two-stage network: the region proposal network (RPN) and the detection head (DH). The RPN generates high-recall proposals, and the DH produces final detection results. The experimental results show the proposed model outperforms more recent research on the KITTI 2D and 3D detection benchmark, particularly for distant and highly occluded instances.
\end{abstract}

\begin{IEEEkeywords}
multi-sensor fusion, autonomous vehicles, vehicle detection, multi-level feature fusion
\end{IEEEkeywords}

\section{Introduction}
Connected and Automated Vehicles (CAVs) are a transformative technology that could significantly change the current transportation modes by replacing human drivers with autonomous controllers. The technology is considered as a mitigator to road accidents and traffic pollution issues \cite{b1, b2}. One of the requirements of implementing autonomous vehicles is safe and collision-free driving, making a trustworthy perception system crucial. The perception system sees the surrounding environment for the autonomous vehicles via a variety of sensors (e.g., Cameras, LIDAR and RADAR) and must accurately detect vehicles, cyclists and pedestrians normally with the aid of the Deep Neural Network.

\begin{figure}[h!]
\centerline{\includegraphics[scale=0.33]{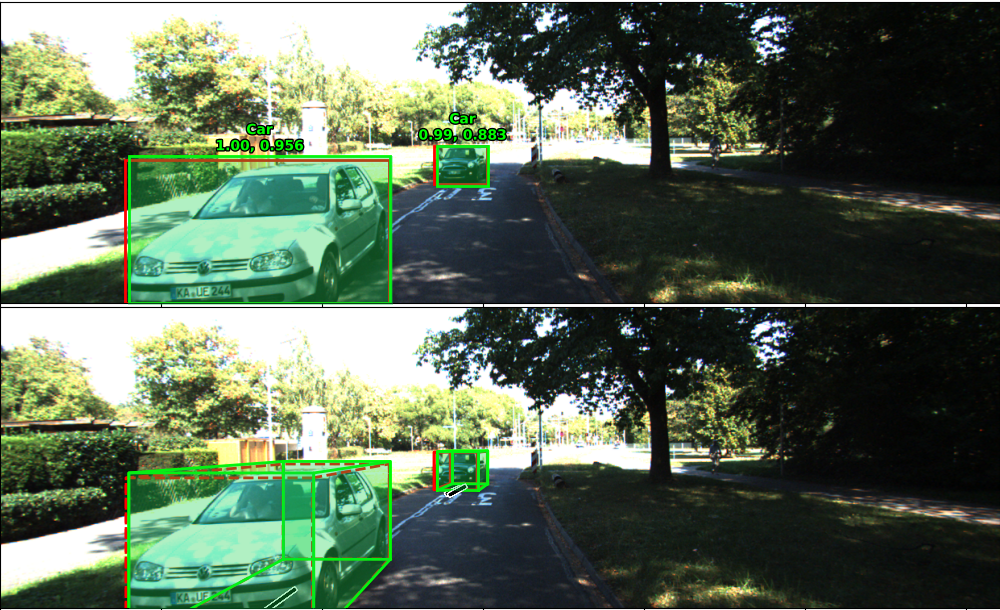}}
\caption{2D and 3D vehicle detection results (green) of the proposed method, tested on the KITTI benchmark \cite{b18}}
\label{fig1}
\end{figure}

The Convolutional Neural Network (CNN) has proven a potent tool to extract and learn features from various sensory modalities. 2D road user detection methods \cite{b3, b4, b5, b6, b7} leverage CNNs to learn features from RGB camera images and have achieved significant and satisfactory results. CNNs can also produce accurate 3D object detection results \cite{b8, b9} by extracting 3D geometry information from the LIDAR point cloud. Some previous work proposes to discretise the 3D point clouds into voxel grids \cite{b10, b11} or preprocess the point cloud into a bird's eye view map \cite{b12} or range view maps \cite{b13} before inputting them to the CNN to mitigate the point cloud's low resolution and sparsity drawbacks. Sensor-fusion-based road user detection methods, which leverage sensors with complementary characteristics, such as combining a LIDAR with a camera or combining a camera with a RADAR, are a hot research topic in improving 3D detection accuracy. This paper focuses on fusing RGB camera images with LIDAR point clouds to enhance 3D detection accuracy by leveraging sufficient RGB information and precise 3D geometry information. As a result, state-of-the-art LIDAR-Camera-fusion-based methods, such as PointPainting \cite{b14}, AVOD \cite{b15}, MVX-Net \cite{b16}, show better performance than some of the SORT LIDAR-based 3D detection methods, e.g. LaserNet \cite{b13} and VoxelNet \cite{b10}. Fusing LIDAR point clouds with RGB images is not a trivial task, and the performance depends on the inputs to the fusion and when the fusion happens. There are three main sensor fusion types, early fusion, middle fusion and late fusion \cite{b17}. However, existing sensor-fusion-based detectors \cite{b12, b16} cannot achieve error-free performance because they cannot fully exploit multi-view features obtained from different levels and the convolutional and downsampling process leads to loss of object features. Loss of details increases the probability of false-negative or false-positive detection and makes detecting partially-occluded or distant vehicles extremely challenging. 

Aiming at these challenges, this paper (Fig.~\ref{fig1}) proposes a novel and comprehensive LIDAR-Camera-fusion-based 3D detection method, which effectively fuses features from both sensory modalities obtained at different levels of the proposed convolutional feature extractor. The architecture of the proposed detection network is illustrated in Fig.~\ref{fig2}. The model takes BEV maps and RGB images as inputs. The novelty of the proposed method is combining a three-stage convolutional neural network (Fig.~\ref{fig3}) with a multi-modal and multi-level fusion scheme (Table 1) to extract the most representative object features across sensory modalities and convolutional layers. These features are shared by a two-stage detection network: the region proposal network (RPN) and the detection head (DH). The RPN projects pre-defined anchors onto feature maps and outputs high-recall 3D proposals. The DH utilises these proposals for the final object classification and oriented bounding box regression.

The proposed approach will be evaluated on the KITTI benchmark \cite{b18} for the 3D vehicle detection evaluation and bird's eye view vehicle detection evaluation. We expect the proposed approach achieves a higher detection accuracy on both evaluation metrics and outperforms some of the more recent 3D detectors, such as VoxelNet \cite{b10}, MV3D \cite{b19}, and MVX-Net \cite{b16}. 

\section{Related Work}
3D object detection research mainly considers RGB sensors, LIDAR, or the fusion of both sensors. 
The majority of existing 3D road user detection methods are LIDAR-based, taking point clouds as input or preprocessing the raw point cloud into a compact representation to improve its sparse, irregular, and borderless characteristics. Classified by the different point cloud representations, the three 3D road user detection methods categories are voxelisation representations, projection-based and point-based.

Point-based methods (e.g., PointNet \cite{b20} and PointNet++ \cite{b21} extract point-wise features directly from the raw LIDAR point cloud without any preprocessing on the raw data.

Voxelisation-based methods \cite{b10} divide the point cloud into equally spaced voxels and extract voxel-wise features through a convolutional neural network. The voxel-wise features are aggregated and passed the region proposal network \cite{b22} for the final object classification and 3D bounding box regression.

Projection-based methods transform the 3D point cloud into 2D representations and apply 2D convolutional neural networks to perform object detection. There are four 2D point cloud representation types: spherical projection (front view map), cylindrical projection, image-plane projection and bird's eye view map projection. VeloFCN \cite{b23} projects the LIDAR point cloud into a 2D cylindrical map, and each point is encoded with its position on the 2D map. Squeezeseg \cite{b24} projects the LIDAR point cloud onto a sphere and characterises each point by azimuth and zenith angles. PointFusion in MVX-Net \cite{b16} applies a pre-trained 2D convolutional network to extract pixel-wise features. Subsequently, it projects the point cloud into the image plane and matches the positions of 3D points with 2D image pixels. According to the position match, each point is appended with the corresponding RGB pixel features. MV3D \cite{b19} proposes the LIDAR BEV maps by discretising the 3D point cloud into a voxel grid, and BEV maps are encoded by points' height, intensity and density. In the proposed approach, we follow MV3D's method to preprocess the 3D point cloud into BEV maps and take it as an input to the detection network. LIDAR BEV maps are capable of mitigating the occlusion problem, enabling the proposed approach to extract highly-detailed features of partially-occluded vehicles.

Image-based 3D detection methods \cite{b25, b26} utilise camera images as input and estimate the 3D bounding box from the 2D bounding box. \cite{b27} proposes a 2D to 3D bounding box estimation by classifying the viewpoint and regressing the centre location projection of the bottom face of 3D bounding boxes. Recently, Traffic-Net \cite{b28}, the state-of-the-art method for traffic monitoring, develops a 2D to 3D model by feature extraction and matching from satellite and ground images.

Sensor-fusion-based 3D detection methods refer to the methods that leverage sensors with complementary characteristics for enhancing detection accuracy under all weather and lighting conditions. PointPainting \cite{b16} and Complexer-YOLO \cite{b29} employ a semantic segmentation network on the RGB images and generate a pixel-wise segmentation mask with scores. The fusion occurs when projecting the point cloud to the mask, and points are decorated according to scores. LaserNet++ \cite{b12} fuses the features extracted from the LIDAR range view map with the RGB image features. Frustum PointNet \cite{b30} generates region proposals from images and then extrudes the region proposals into 3D frustum region proposals. Then they apply a semantic segmentation network to filter the points, and the rest of the points are processed by the PointNet \cite{b20} for bounding box regression.

MV3D \cite{b19}, a two-stage detection method, generates proposals based on the features extracted from the LIDAR BEV maps. Then, it uses these proposals to fuse element-wise features from LIDAR BEV maps, LIDAR range view maps and RGB images. However, points on LIDAR BEV maps are sparse, which is less suitable for detecting small or partially-occluded or distant road users. Therefore, MV3D's RPN is limited from generating reliable proposals for small, distant or partially-occluded instances. Although it fuses multi-features in the detection stage, the performance of MV3D is severely limited by the performance of its RPN. 

\section{Methodology}
\begin{figure*}[h!]
\centerline{\includegraphics[scale=0.33]{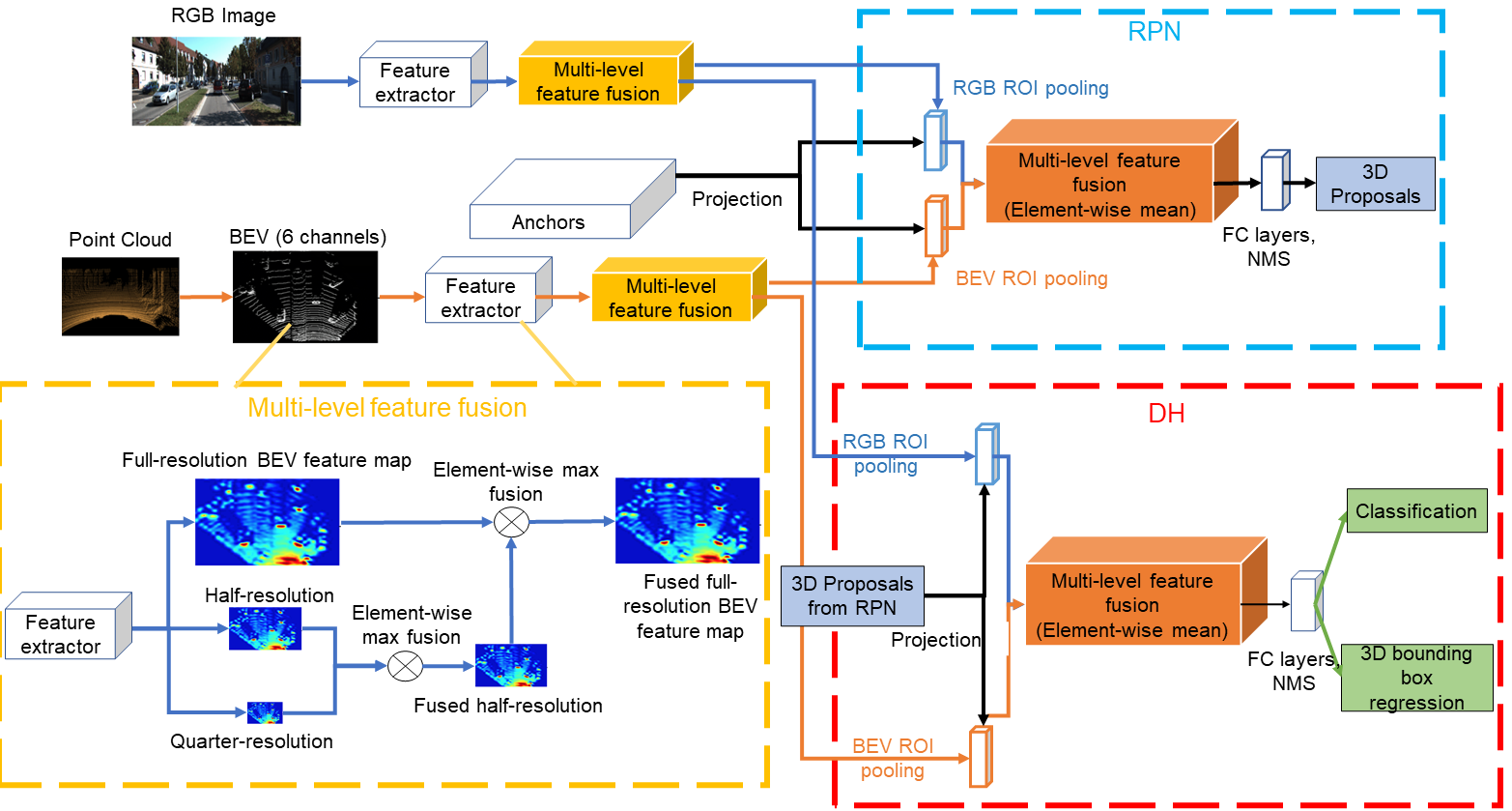}}
\caption{The architecture of the proposed detection network. Yellow boxes: multi-level feature fusion. Orange boxes: multi-modal feature fusion. The blue dotted box: the region proposal network (RPN). The red dotted box: the detection head (DH).}
\label{fig2}
\end{figure*}

\begin{figure}[h!]
\centerline{\includegraphics[scale=0.25]{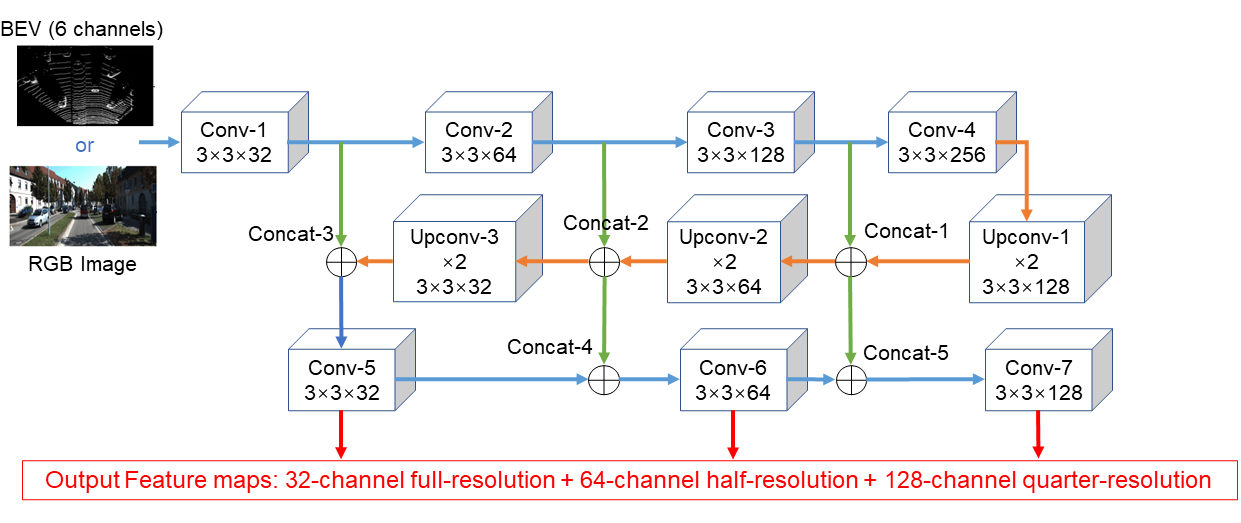}}
\caption{The architecture of the proposed three-stage feature extractor contains two convolutional stages and one deconvolutional stage. The lateral connections are shown in green lines.}
\label{fig3}
\end{figure}

This paper proposes a two-stage 3D detector to achieve high-accuracy detection of distant and partially-occluded vehicles. As illustrated in Fig.~\ref{fig2}, the proposed detection model takes RGB images and LIDAR point clouds as inputs. The model first generates LIDAR BEV maps from the LIDAR point cloud. The novelty of the proposed method is employing a three-stage feature extractor and a novel sensor fusion scheme to extract the most representative features from two input views effectively and enable the recovery of details discarded during the downsampling process. A two-stage network shares the extracted features: the 3D region proposal network (RPN) and the detection head (DH) module. The RPN generates high-recall 3D proposals by projecting pre-defined anchors onto the feature maps and employing feature pooling. Finally, the DH module utilises these proposals for the final object classification and oriented 3D bounding box regression. The detailed object features extracted by the feature extractor and the sensor fusion scheme increases the model's accuracy in detecting partially occluded and distant vehicles.

\subsection{LIDAR BEV maps generator}\label{AA}
Six-channels LIDAR BEV maps are generated from the point cloud following the method proposed in MV3D \cite{b19}, and the BEV maps are encoded by the height and density of points. The point cloud is first divided equally into five slices along the z-axis and then projected into a 2D grid with a resolution of 0.1m. Each slice generates one height map by encoding each grid cell’s maximum height of points. Five height channels are generated for one point cloud input. The sixth channel of the BEV maps is the density map, which encodes the density of points in each cell, and is computed as shown in equation (1) for each cell, where N is the number of points in each cell.
\begin{equation}
d = min(1.0,\frac{log(N+1)}{log(64)})
\end{equation}

\begin{table*}[h!]
\footnotesize
\footnotesize
\renewcommand\arraystretch{1.2}
\caption{The architectures of the proposed multi-level and multi-modal sensor fusion scheme.}
\begin{center}
\resizebox{0.72\linewidth}{!}{\begin{tabular}{|c|c|c|c|c|}
\hline
\multirow{6}{*}{\textbf{Multi-level Fusion}} & Bilinear-1 & 2× & 180×600×128 & 352×400×128 \\\cline{2-5}
                                    & Conv-8 & 3×3×64 & 180×600×64 & 352×400×64 \\\cline{2-5}
                                    & Element-wise-Max-1 & Conv-8+Conv-6 & 180×600×64 & 352×400×64 \\\cline{2-5}
                                    & Bilinear-2 & 2× & 360×1200×64 & 704×800×64 \\\cline{2-5}
                                    & Conv-9 & 3×3×32 & 360×1200×32 & 704×800×32 \\\cline{2-5}
                                    & Element-wise-Max-2 & Conv-9+Conv-5 & 360×1200×32 & 704×800×32\\\hline
\textbf{Dimensionality reduction} & Conv-10 & 1×1×1 & 360×1200×1 & 704×800×1\\
\hline
\multirow{2}{*}{\textbf{Multi-modal Fusion}} & \multicolumn{4}{|c|}{Anchor/Proposal Projection + Feature Pooling}\\\cline{2-5}
                                             & \multicolumn{4}{|c|}{Element-wise mean fusion (Image feature vector + BEV feature vector)}\\\hline
\end{tabular}}
\label{table1}
\end{center}
\vspace{-2mm}
\end{table*}

\subsection{Feature extractor}
Inspired by the powerful 2D feature extractor proposed in U-Net \cite{b31}, a three-stage feature extractor (Fig.~\ref{fig3}) is proposed to extract high-level object features from RGB images and BEV maps, including two convolutional stages and one deconvolutional stage. Given the high performance of VGG-16 \cite{b4} in extracting object features, the first convolutional stage of the proposed feature extractor is built on modifying VGG-16’s neural network architecture, and the modifications include reducing the number of channels and dropping off the fifth set of layers. Reducing the number of channels can mitigate the computation and memory overhead. The fifth set of layers is discarded because hard-difficulty instances are struggling to be accurately detected on feature maps after being downsampled by sixteen times. In addition, inaccurate hard-difficulty object features are harmful to the details recovery process in the following two stages, leading to accuracy drops. Therefore, the first convolutional stage finally outputs a high-level feature map with one-eighth resolution. 

To recover object features from low resolution to full resolution, the proposed deconvolutional stage utilises transposed layers (deconvolutional layers) to achieve trainable upsampling. The one-eighth-resolution feature maps from the first convolutional stage are gradually upsampled to full resolution through three transposed layers. Compared with bilinear upsampling, which utilises linear interpolations to calculate the pixels’ values from the nearby pixels, transposed layers provide a more trainable and effective upsampling process. 

However, the details lost in the downsampling and convolutional process cannot be recovered by the deconvolutional layers, and inaccurate features are harmful to the convolutional process, increasing the probability of False-Positive detection. To this end, lateral connections are applied to concatenate feature maps of the same spatial sizes across three stages. As shown in Fig.~\ref{fig3}, the outputs from the first convolutional stage are first upsampled by two times via the deconvolutional layer Upconv-1, and the outputs are concatenated with the feature maps from Conv-3 layers in the first stage along the channel dimensionalities. Then, Upconv-2, the next deconvolution layer, takes these concatenated feature maps as inputs and outputs double-sized feature maps after selecting more representative object features from feature maps of the same spatial sizes from different stages. Finally, the double-sized feature maps are concatenated with the ones from Conv-2 layers and upsampled to the full resolution via a deconvolutional layer. The gradual upsampling and lateral connections in this stage offer the deep neural network a chance to re-select more representative object features globally and discard inaccurate features by setting different weights to channels of feature maps. 

The second convolutional stage is designed to enhance features’ qualities and generate higher-level object features. Since features of distant or partially occluded objects may occupy less than one pixel on the eighth-resolution feature map, only three sets of convolutional layers and two max-pooling layers are employed in this stage. Lateral connections are employed again to retain features of distant or partially-occluded objects in the final outputs of the feature extractor. The outputs from the deconvolutional stage are first concatenated with feature maps from Conv-1 layers along channel dimensionalities. The concatenated features are processed by Conv-5 layers and downsampled by a max-pooling layer. The downsampled feature maps are concatenated with those from Conv-2 and Upconv-2 layers, and Conv-6 layers select accurate object features and discard inaccurate ones from concatenated feature maps. Finally, the outputs from Conv-6 layers are concatenated with those from Conv-3 and Upconv-1 layers and processed by Conv-7 layers. The final output of the feature extractor includes one full-resolution feature map, one half-resolution and one quarter-resolution.

To be noted, the features from the first stage are concatenated to the later stages twice to enhance the performance of features recovery because these features are generated at an early period and preserve the initial and accurate object and background low-level information, making them essential to support achieving details recovery when high-level object features coexist with inaccurate features at a later period. 

\subsection{Multi-modal and Multi-level feature pooling and fusion}
The proposed approach projects 3D anchors and 3D proposals onto multi-view feature maps from the feature extractor to obtain region-wise features. Then the proposed method applies a novel multi-level and multi-modal fusion scheme to fuse these region-wise features across different convolutional layers and data types, enabling the model to find the most representative features element-wise.

The proposed sensor fusion scheme (Fig.~\ref{fig2} and Table~\ref{table1}) contains four blocks: multi-level feature fusion, dimensionality reduction, feature pooling and multi-modal feature fusion, and it finally outputs element-wise fused feature vectors to fully connected layers.

The multi-level fusion scheme is designed to achieve feature fusion across convolutional layers for each input view and generate a full-resolution feature map containing the most representative object features. It takes the outputs from the feature extractor as inputs and hierarchically fuses them pixel-wise. 2× bilinear upsampling is first applied to upsample the quarter-resolution feature map to half-resolution, and the channel dimensionalities are reduced from 128 to 64 via Conv-8 layers. A bilinear upsampling operation is selected in this fusion scheme because it shows relatively stable performance and lower computational costs than the nearest-neighbour, bicubic and deconvolution methods. Afterwards, the upsampled feature map is fused with the half-resolution feature map from the feature extractor along channels by applying the element-wise max operation to extract the more representative pixel-wise features. Subsequently, a 2× bilinear upsampling is applied on the fused feature map to raise the size to full resolution, and the channel dimensionality is reduced from 64 to 32 through Conv-9 layers. Finally, this new feature map and the full-resolution feature map from the feature extractor are fused by applying the element-wise max operation to find the most representative pixel-wise features across different convolutional layers. The fused full-resolution feature map, containing pixel-wise maxima, is the output of the multi-level fusion scheme, representing the most representative object features extracted from each input view.

In the RPN module, a 1×1×1 convolutional layer is applied after the multi-level fusion to reduce the channel dimensionality from thirty-two to one. This dimensionality reduction fuses the pixel-wise features across channels onto one feature map and reduces the memory and computational overhead.

3D anchors are projected onto each view’s one-channel full resolution feature map, and they are transformed into 2D regions of interest (ROIs) in the RPN. The projection and transformation are achieved by transforming the LIDAR coordinates into BEV and image coordinates. The KITTI dataset \cite{b18}, used for training and testing the proposed model, provides the LIDAR coordinate system and the calibration matrix to support the transformation and projection between points to pixels. ROI pooling is employed inside each ROI to obtain a fixed-length feature vector element-wise. 

The element-wise mean operation is employed to fuse feature vectors from two input views, assigning both views with equal weights, and the output of the proposed feature fusion scheme is a fused feature vector element-wise.

\subsection{3D region proposal network (RPN)}
Similar to the Region Proposal Network \cite{b22}, a set of 3D anchors based on the KITTI dataset \cite{b18} is pre-designed and parametrised by (x, y, z, l, w, h). (x, y, z) refers to the centre coordinates of an anchor. z is computed as the distance from the LIDAR to the ground. x and y are varying positions sampled in the point cloud along the x-axis and y-axis at an interval of 0.5 meters. (l, w, h) refer to the dimensions of an anchor box, and they are computed by clustering the samples of each class in the KITTI training set. The proposed approach computed two sets of length and width values for the car class. By rotating each anchor box 90 degrees, four anchors are centred at an (x, y) point. The anchor boxes on the BEV map are parameterised as ($x_{bev}$, $y_{bev}$, $l_{bev}$, $w_{bev}$) by discretising (x, y, l, w). The anchor boxes with no points inside are dropped. The rest of the anchors are projected onto multi-view feature maps, and we employ our multi-modal and multi-level feature pooling and fusion scheme to obtain a fused feature vector for each region of interest. The fused feature vector is fed to fully connected layers for the probability score (softmax output) and 3D bounding box regression tasks. Orientation regression is not considered in the RPN module, and the boxes only head towards 0 and 90 degrees. The model employs a binary cross-entropy loss function, which compares the predicted probabilities to the ground truth to measure the classification loss. The loss increases when the predicted probability becomes far from the ground truth. The 3D bounding box is regressed by computing ($\Delta$x, $\Delta$y, $\Delta$z, $\Delta$l, $\Delta$w, $\Delta$h), and the model employs the Smooth L1 loss function proposed in Fast-CNN \cite{b32} to compute the regression loss. Smooth L1 loss minimises the sum of the absolute values of differences between the ground truth and the predicted value and shows negligible sensitivity to outliers. In addition, ($\Delta$x, $\Delta$y) are normalised with the diagonal d and ($\Delta$z) is normalised with the anchor’s height.

\begin{equation}
d = \sqrt{{(l_{anchor)}}^2+{(w_{anchor)}}^2}
\end{equation}

\begin{equation}
\Delta{x}, \Delta{y} = \frac{x,y_{GT}-x,y_{anchor}}{d}
\end{equation}

\begin{equation}
\Delta{z} = \frac{z_{GT}-z_{anchor}}{h_{anchor}}
\end{equation}

\begin{equation}
\Delta{l}, \Delta{w}, \Delta{h} = log(\frac{l,w,h_{GT}}{l,w,h_{anchor}})  
\end{equation}
Where $l, w, h_{anchor}$ are anchor's length, width and height, $l, w, h_{GT}$ are ground truth box's length, width and height, $x, y, z_{GT}$ are centre coordinates of the ground truth box, and $x, y, z_{anchor}$ are centre coordinates of the anchor.

Like the RPN \cite{b22}, 2D IoU on the BEV map is computed between anchors and ground truth boxes, and background anchors are ignored when computing the regression loss. IoU threshold is set to 0.4 and 0.6 in terms of the car class. If IoU is above 0.6, anchors are positive. If IoU is below 0.4, anchors are negative. 2D non-maximal suppression (NMS) is applied to the predicted boxes to suppress redundant boxes, with an IoU threshold of 0.7.

\subsection{Detection Head (DH)}
3D proposals generated by the RPN are projected onto 32-channel full-resolution feature maps from the feature extractor. Dimensionality reduction is not implemented in the DH module, considering that the number of proposals is much lower than anchors. Following the same methods taken in the RPN module, a fused feature vector is obtained for each ROI, and the fully connected layers process the fused feature vectors for the final classification and 3D bounding box regression tasks. 

To regress the orientation of the 3D bounding box, we parametrise a 3D ground truth box as (x, y, z, l, w, h, $\theta$), where $\theta$ is the yaw rotation angle. The orientation of the 3D bounding box is estimated directly. Similar to the methods taken in the RPN module, the model applies the binary cross-entropy loss function to compute the classification loss and the SmoothL1 loss function \cite{b32} to compute the regression loss of ($\Delta$x, $\Delta$y, $\Delta$z, $\Delta$l, $\Delta$w, $\Delta$h, $\Delta$$\theta$). Regression of ($\Delta$x, $\Delta$y, $\Delta$z, $\Delta$l, $\Delta$w, $\Delta$h) follows the same equations utilised in the RPN module, and the regression of orientation $\theta$ is computed as:
\begin{equation}
\Delta{\theta} = \theta_{GT}-\theta_{anchor}
\end{equation}
Where $\theta_{GT}$ and $\theta_{anchor}$ are orientation angles of the ground truth box and the anchor, respectively.

The model computes 2D IoU between proposals and ground truth boxes during training. The IoU threshold is set to 0.7 for the car class following the same instructions taken in Pyramid R-CNN \cite{b18, b33}. The 3D proposals that have IoU above the threshold are considered positive, and proposals are considered negative when their IoU is below the threshold. Only positive proposals are taken into computing the regression loss. NMS is applied to remove redundant boxes, with an IoU threshold of 0.01.

\section{Experiments and Results}
The experiment focuses on vehicle detection. The model is trained and tested on the challenging KITTI benchmark \cite{b18}, containing 7481 training images, 7518 test images, as well as the corresponding point cloud of  each image. The dataset includes 80256 labelled objects, covering three classes: car, pedestrian and cyclists. For each class, there are three difficulty levels of objects: easy, moderate and hard. They are determined based on the object's size, occlusion level, visibility and truncation level. The KITTI dataset also provides the calibration matrix, which contains the spatial information between image pixels and 3D point cloud coordinates. As the KITTI test set does not provide ground truth boxes and labels, we divide the training set into two halves, training and validation sets, following the method proposed by \cite{b34, b35}. The training set is used for the training purpose, and the proposed approach is evaluated on the validation set to compare the performance with other SORT 3D detection methods.

In the training phase, we applied the adaptive moment estimation (Adam) optimiser to update network weights, iteratively. Adam optimiser utilises the gradient descent with momentum and the Root Mean Square propagation algorithms and shows substantial advantages in efficient computation with low memory requirements. The proposed approach is trained for 120k iterations in total, which is the same total iterations used in the sensor-fusion-based 3D detection network, MV3D \cite{b19}. The initial learning rate is set as 0.0001 and exponentially decays with a factor of 0.8 at every 20k iteration, which is slightly modified based on the parameters used in MV3D \cite{b19}. This training strategy gives an intuitive comparison of performance between the proposed model and other similar research work. Specifically, we can directly compare the sensor fusion method and the feature extraction method with the methods used in MV3D, when two models take the same inputs and employ similar training strategies.

We cluster the car class samples in the training set to compute two sets of length and width values for the car class, and these values are used for pre-defining the anchors. The height of the anchor is set to 1.65m, which is the camera height above the ground. The anchors are pre-generated above the ground plane for each RGB image in the training and validation sets based on the ground plane coefficients $(ax + by + cz + d) = 0$.

Finally, the proposed approach is tested on the KITTI validation set by setting the IoU threshold to 0.7 for the car class, and its 3D detection and BEV detection performance are compared with four more recent and extensively tested 3D detection methods: MV3D \cite{b19}, VoxelNet \cite{b10}, MVX-Net \cite{b16} and F-PointNet \cite{b30}, shown in Table~\ref{table2} and ~\ref{table3}.

The BEV detection accuracy evaluation result is illustrated in Table~\ref{table2}. The proposed approach outperforms the other four 3D detection methods in detecting easy and moderate vehicles. It achieves the second-best accuracy in detecting hard vehicles. It is noted that our approach outperforms the two-stage sensor-fusion-based model MV3D \cite{b19} by 8.89\% in detecting moderate vehicles and 3.03\% in detecting hard vehicles. The proposed approach also outperforms the LIDAR-Point-Cloud-based method VoxelNet \cite{b10} and achieves a 1.88\% accuracy lead in detecting moderate vehicles. The success shows the proposed feature extractor and the proposed sensor fusion scheme can effectively extract and generate high-level object features from both input views and recover distant and partially-occluded vehicle features in the final output. The results also show the multi-level texture information from image planes can be efficiently fused with the multi-level spatial information from LIDAR BEV maps, leading to considerable enhancement in BEV detection accuracy.

\begin{table}[t!]
\footnotesize
\caption{Comparison of the proposed model with four 3D detection methods in terms of the BEV vehicle detection accuracy, evaluated on the KITTI validation set}
\begin{center}
\begin{tabular}{|c|c|c|c|}
\hline
\multirow{2}{*}{\textit{Method}} &
\multicolumn{3}{|c|}{Car $AP_{BEV} (\%)$}\\\cline{2-4} & Easy & Moderate & Hard \\\hline
\textit{MV3D \cite{b19}} & 86.55 & 78.10 & 76.77 \\\hline
\textit{VoxelNet \cite{b10}} & 89.60 & 84.81 & 78.57 \\\hline
\textit{MVX-Next \cite{b16}} & 88.60 & 84.60 & 78.60 \\\hline
\textit{F-PointNet \cite{b30}} & N/A & N/A & N/A \\\hline
\textit{Ours} & \textbf{89.62} & \textbf{89.69} & 79.80 \\\hline
\end{tabular}
\label{table2}
\end{center}
\end{table}

\begin{table}[t!]
\footnotesize
\caption{Comparison of the proposed model with four 3D detection methods in terms of the 3D vehicle detection accuracy, evaluated on the KITTI validation set}
\begin{center}
\begin{tabular}{|c|c|c|c|}
\hline
\multirow{2}{*}{\textit{Method}} &
\multicolumn{3}{|c|}{Car $AP_{3D} (\%)$}\\\cline{2-4} & Easy & Moderate & Hard \\\hline
\textit{MV3D \cite{b19}} & 71.29 & 62.68 & 56.56 \\\hline
\textit{VoxelNet \cite{b10}} & 81.79 & 65.46 & 62.85 \\\hline
\textit{MVX-Next \cite{b16}} & 82.30 & 72.20 & 66.80 \\\hline
\textit{F-PointNet \cite{b30}} & 83.76 & 70.92 & 63.65 \\\hline
\textit{Ours} & 83.27 & \textbf{73.99} & \textbf{67.92} \\\hline
\end{tabular}
\label{table3}
\end{center}
\end{table}

\begin{figure}[h!]
\centerline{\includegraphics[scale=0.28]{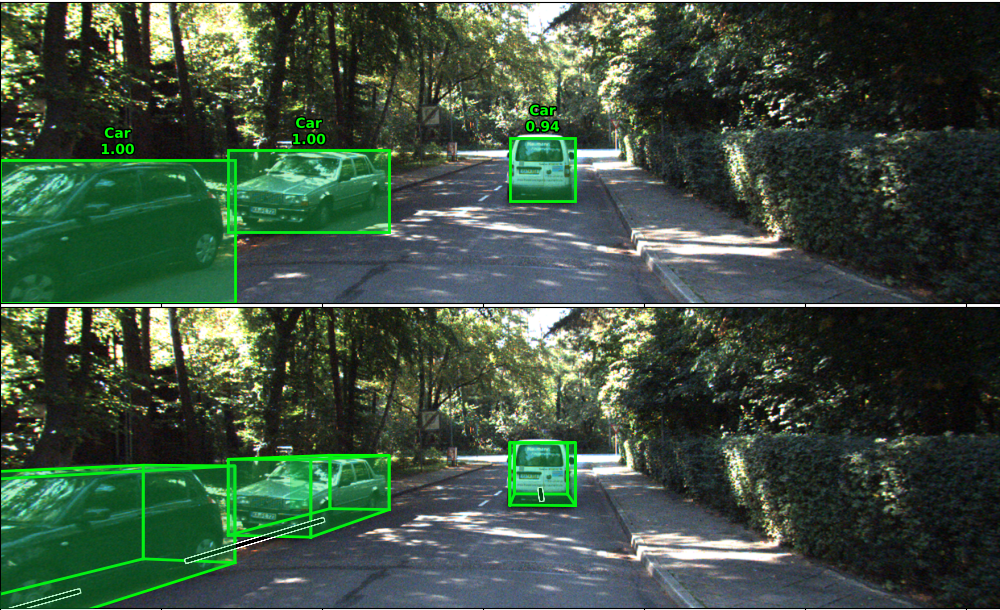}}
\caption{2D and 3D vehicle detection results (green) of the proposed method in a standard driving scenario.}
\label{fig4}
\end{figure}

\begin{figure}[h!]
\centerline{\includegraphics[scale=0.28]{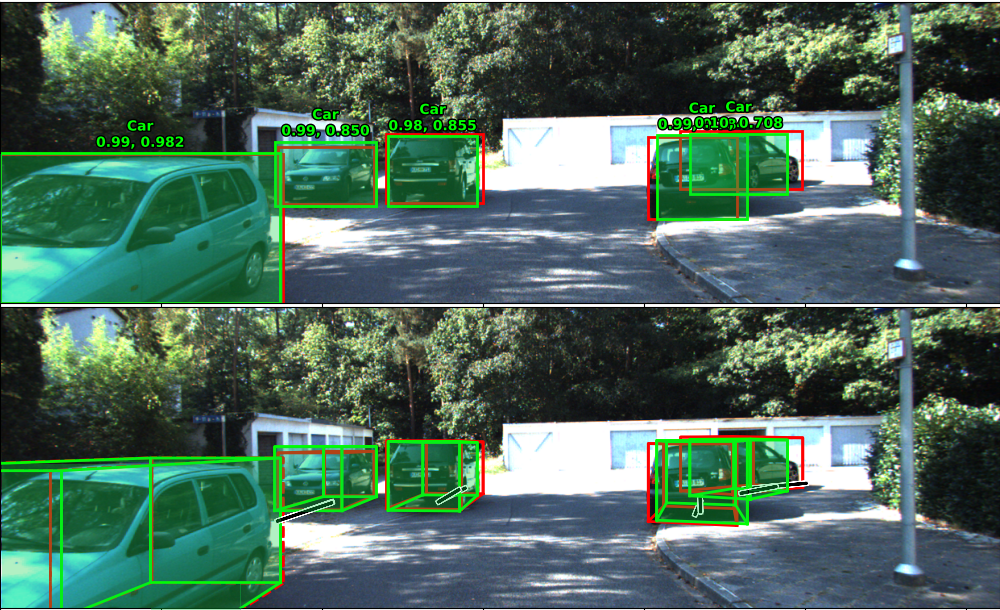}}
\caption{2D and 3D vehicle detection results (green) of the proposed method in a partially-occluded driving scenario.}
\label{fig5}
\end{figure}

\begin{figure}[h!]
\centerline{\includegraphics[scale=0.28]{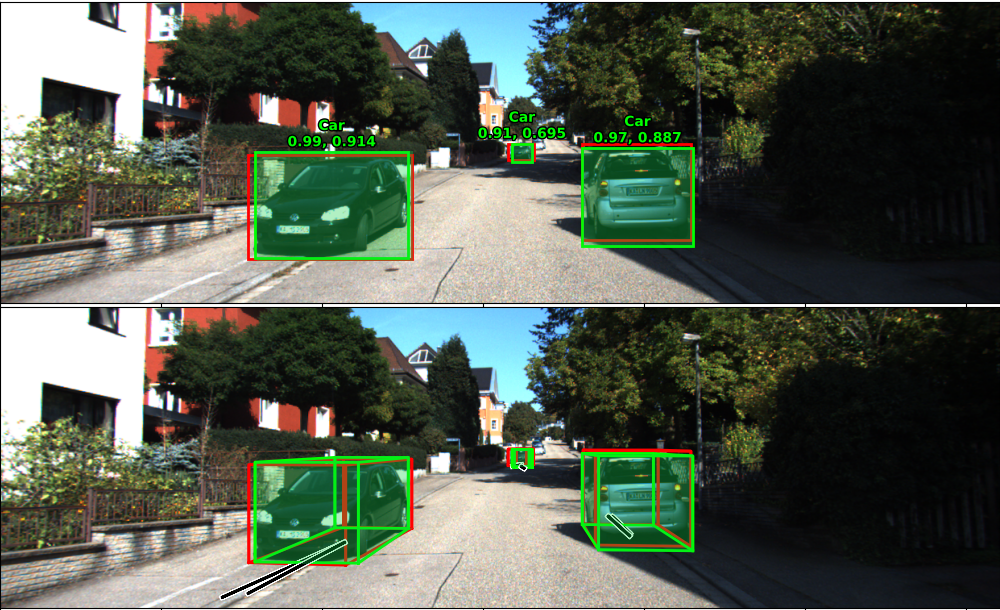}}
\caption{2D and 3D vehicle detection results (green) of the proposed method in a driving scenario with distant vehicles.}
\label{fig6}
\end{figure}

The evaluation results on the 3D vehicle detection accuracy are presented in Table~\ref{table3}. Although the proposed approach achieves the second-best 3D detection accuracy in detecting easy vehicles, it achieves the best performance in 3D detection of moderate and hard vehicles. The proposed approach outperforms the MV3D \cite{b19} by 11.89\%, 11.31\% and 11.36\% accuracy in detecting easy, moderate and hard vehicles, respectively. The notable leads in detection performance prove the significance of interactions between features from different stages of convolution layers and different sensory modalities. Compared the results with the LIDAR-solely-based VoxelNet \cite{b10}, the proposed approach achieves a significant lead of 8.53\% in detecting moderate vehicles. It is also noted that the proposed approach outperforms the state-of-the-art sensor-fusion-based method MVX-Net \cite{b16} in detecting moderate and hard vehicles. The results prove the noteworthy improvements achieved by the proposed method in enhancing the detection accuracy of partially-occluded or distant vehicles. This improvement further proves that the proposed approach efficiently and effectively fuse images' texture information and point cloud's spatial information, by taking advantage of each modality and mitigating their drawbacks.
Some 2D and 3D detection results obtained in different driving scenarios can be visualised in Figure~\ref{fig4} to~\ref{fig6}. The figure~\ref{fig4} shows that the proposed model can accurately detect all vehicles in a standard driving scenario. Figure~\ref{fig5} shows that the proposed approach can accurately detect all vehicles in a congested traffic scenario, including partially-occluded and low-visibility vehicles. In Figure~\ref{fig6}, it can be visualised that the proposed approach can accurately detect distant vehicles, which appear small on the image plane.

\section{Conclusion}
This article proposed a two-stage object detector for driving scenarios that applies the convolutional neural network to extract features from RGB images and LIDAR BEV maps. A multi-modal and multi-level feature fusion scheme was developed to fuse high-level object features across input views and convolutional layers. It achieved taking advantage of each sensor and mitigating sensors' drawbacks with complementary characteristics. A three-stage feature extractor, including two convolutional stages and one deconvolutional stage, was 
designed to extract high-level object features from multi-views for the feature fusion scheme. The feature extractor employed lateral connections between stages to support the details recovery. The designed feature extractor and sensor fusion scheme led to an 11.36\% improvement in 3D KITTI hard vehicles detection accuracy compared with the two-stage detector MV3D \cite{b19}, showing its satisfactory performance in detecting distant, partially occluded and low-visibility instances. 

The proposed approach was tested on the KITTI validation set and the 3D and BEV vehicle detection outperformed four more recent 3D detection methods, including MV3D \cite{b19}, MVX-Net \cite{b16}, F-PointNet \cite{b30} and VoxelNet \cite{b10}. It is noted that the proposed model outperformed the MVX-Net \cite{b16} and F-PointNet \cite{b30} in 3D detection of KITTI’s moderate and hard vehicles. This performance lead can be attributed to employing the proposed three-stage feature extractor and the feature fusion scheme. The drawbacks of the proposed model, which prevent it from achieving higher accuracy, can be categorised into three causes: BEV maps, anchors, and ground planes. Specifically, the LIDAR BEV maps can 
mitigate the occlusion issue but suffer from the sparsity of points. The dimensions of anchors 
were pre-defined by clustering objects in the training set, preventing the RPN from detecting
objects that have considerably larger or smaller spatial sizes. In addition, the proposed approach requires a highly accurate ground plane estimation method to measure ground planes for pre-defining anchors, which is inefficient when implemented in the real world.

Aiming toward higher detection accuracy, we may suggest increasing BEV channels as a future work by generating more height maps to mitigate the sparsity of points on BEV maps. The availability of a larger-scale dataset can further increase detection accuracy by providing more training labels in various driving scenarios. 
\pagebreak

\end{document}